\documentclass[sigconf,nonacm]{acmart}

\usepackage[utf8]{inputenc} 
\usepackage[T1]{fontenc}    
\usepackage{hyperref}       
\usepackage{booktabs}       
\usepackage{amsfonts}       
\usepackage{nicefrac}       
\usepackage{microtype}      
\usepackage{xcolor}         
\usepackage{multicol}

\usepackage{amsmath,wasysym}
\usepackage[ruled,vlined,linesnumbered, resetcount]{algorithm2e}

\usepackage{subcaption}
\usepackage{multirow}
\usepackage{textcomp}
\usepackage{footnote}
\usepackage{makecell}
\usepackage{xspace}
\usepackage{enumitem}
\usepackage{wrapfig}
\usepackage{amsthm}
\usepackage{wrapfig}

\newcommand{\model}{\textsc{StoIC}\xspace}
\newcommand{\pte}{\textsc{PTE}\xspace}
\newcommand{\scg}{\textsc{RCN}\xspace}

\newcommand{\ggm}{\textsc{GGM}\xspace}
\newcommand{\rgne}{\textsc{RGNE}\xspace}

\newcommand{\gts}{\textsc{GTS}\xspace}
\newcommand{\nir}{\textsc{NRI}\xspace}
\newcommand{\mtgnn}{\textsc{MTGNN}\xspace}
\newcommand{\deepar}{\textsc{DeepAR}\xspace}
\newcommand{\tsfnp}{\textsc{EpiFNP}\xspace}
\newcommand{\gdn}{\textsc{GDN}\xspace}
\newcommand{\lds}{\textsc{LDS}\xspace}

\newcommand{\modelng}{\textsc{StoIC-NoGraph}\xspace}
\newcommand{\modelns}{\textsc{StoIC-NoRefCorr}\xspace}
\newcommand{\modelnc}{\textsc{StoIC-NoWtAgg}\xspace}

\newcommand{\flu}{\texttt{Flu-Symptoms}\xspace}
\newcommand{\covid}{\texttt{Covid-19}\xspace}
\newcommand{\stocks}{\texttt{S\&P-100}\xspace}
\newcommand{\power}{\texttt{Electricity}\xspace}
\newcommand{\trafficla}{\texttt{METR-LA}\xspace}
\newcommand{\trafficbay}{\texttt{PEMS-BAYS}\xspace}
\newcommand{\nycb}{\texttt{NYC-Bike}\xspace}
\newcommand{\nyct}{\texttt{NYC-Taxi}\xspace}

\begin{document}

\title{Learning Graph Structures and Uncertainty for Accurate and Calibrated Time-series Forecasting}

\author{%
  Harshavardhan Kamarthi}
  \affiliation{
  College of Computing, 
  Georgia Institute of Technology
  \country{USA}}
  \email{hkamarthi3@gatech.edu}
  \author{%
  Lingkai Kong}
  \affiliation{
  College of Computing, 
  Georgia Institute of Technology
  \country{USA}}
  \email{lkkong@gatech.edu}
  \author{%
  Alexander Rodr\'iguez}
  \affiliation{
  College of Computing, 
  Georgia Institute of Technology
  \country{USA}}
  \email{arodriguezc@gatech.edu}
  \author{%
  Chao Zhang}
  \affiliation{
  College of Computing, 
  Georgia Institute of Technology
  \country{USA}}
  \email{chaozhang@gatech.edu}
  \author{%
  B. Aditya Prakash}
  \affiliation{
  College of Computing, 
  Georgia Institute of Technology
  \country{USA}}
  \email{badityap@cc.gatech.edu}

\begin{abstract}
  Multi-variate time series forecasting is an important problem with a wide range of applications.
Recent works model the relations between time-series as graphs and have shown that propagating information over the relation graph can improve time series forecasting.
However, in many cases, relational information is not available or is noisy and reliable.
Moreover, most works ignore the underlying uncertainty of time-series both for structure learning and
deriving the forecasts resulting in the structure not capturing the uncertainty resulting in forecast distributions with poor uncertainty estimates.
We tackle this challenge and introduce \model, that leverages stochastic correlations between
time-series to learn underlying structure between time-series and
to provide well-calibrated and accurate forecasts.
Over a wide-range of benchmark datasets \model provides around 16\% more accurate and 14\% better-calibrated forecasts.
 \model also shows better adaptation to noise in data during inference
and captures important and useful relational information in various benchmarks.

\end{abstract}


\maketitle

\section{Introduction}
\label{sec:intro}

While there has been a lot of work on modeling and forecasting univariate time-series \cite{hyndman2018forecasting},
the problem of multivariate time-series forecasting is more challenging.
This is because modeling individual signals independently may not be sufficient to capture the underlying relationships
between the signals which are essential for strong predictive performance.
Therefore, many multivariate models model sparse correlations between signals based on prior knowledge of underlying structure using Convolutional networks \cite{li2017diffusion}
or Graph Neural networks \cite{kipf2016semi,zhao2019t,yu2017spatio}.

However, in many real-world applications, the graph structure is not available or is unreliable.
In such cases, the problem of learning underlying patterns \cite{franceschi2019learning,shang2021discrete,pamfil2020dynotears}
is an active area of research \cite{zugner2021study} in applications
such as traffic prediction and energy forecasting.
Most methods use a joint learning approach to train the parameters of both graph inference and forecasting modules.

However, most previous works focus only on point forecasting and do not leverage uncertainty when modeling the structure.
Systematically modeling this uncertainty into the modeling pipeline can   help the model adapt to unseen patterns such as when modeling a novel pandemic \cite{rodriguez2021deepcovid}.
Therefore, the learned structure from existing models may not be adapted to noise in data or to distributional shifts commonly encountered in real-world datasets.

In this paper, we tackle the challenge of leveraging structure learning to provide \textit{accurate} and \textit{calibrated} probabilistic forecasts
for all signals of a multivariate time-series.
We introduce a novel probabilistic neural multivariate time-series model,
\model (\underline{Sto}chastic Graph \underline{I}nference for \underline{C}alibrated Forecasting),
that leverages functional neural process framework \cite{louizos2019functional} to model uncertainty in temporal patterns
of individual time-series as well as a joint structure learning module that leverages both pair-wise similarities of time-series
and their uncertainty to model the graph distribution of the underlying structure.
\model then leverages the distribution of learned structure to provide accurate and calibrated forecast distributions
for all the time-series.

Our contributions can be summarized as follows:
(1) \textbf{Deep probabilistic multivariate forecasting model using Joint Structure learning:}
          We propose a Neural Process based probabilistic deep learning model that captures complex temporal
          and structural correlations and uncertainty over multivariate time-series. 
    (2) \textbf{State-of-art accuracy and calibration in multivariate forecasting:} We evaluate \model against
          previous state-of-art models in a wide range of benchmarks
          and observe 16.5\% more accurate and 14.7\% better calibration performance.
          We also show that \model
          is significantly better adapted to provide consistent performance with the injection of varying degrees of noise into datasets due to modeling uncertainty.
    (3) \textbf{Mining useful structural patterns:} We provide multiple case studies to show that \model
          identifies useful domain-specific patterns based on the graphs learned such as modeling
          relations between stocks of the same sectors, location proximity in traffic sensors, and epidemic forecasting.

\section{Methodology}
\subsubsection{Problem Formulation}Consider a multi-variate time-series dataset $\mathcal{D}$ of $N$ time-series $\mathcal{D} =\{\mathbf{y}_i\}_{i=1}^N$ over $T$ time-steps. 
Let $\mathbf{y}_i \in \mathbb{R}^T$ denote time-series $i$ and $y_i^{(t)}$ be the value at time $t$. 
Further, let $\mathbf{y}^{(t)} \in \mathbb{R}^N$ be the vector of all time-series values at time $t$. 
Given time-series values from till current time $t$ as $\mathbf{y}^{(1:t)}$, the goal of probabilistic multivariate forecasting is to train a model $M$ that provides a forecast distribution:
\begin{equation}
    p_M\left( \mathbf{y}^{(t+1:t+\tau)} | \mathbf{y}^{(1:t)}; \theta \right),
\end{equation}
which should be \textit{accurate}, i.e, has mean close to ground truth as well as \textit{calibrated}, i.e., the confidence intervals of the forecasts precisely
mimic actual empirical probabilities \cite{fisch2022calibrated, kamarthi2021doubt}.

Formally,
the goal of joint-structure learning for probabilistic forecasting is to learn a global graph $G$ from $\mathbf{y}^{(1:t)}$ and leverage it to provide \textit{accurate} and \textit{well-calibrated} forecast distributions:
\begin{equation}
    p_M\left( \mathbf{y}^{(t+1:t+\tau)} | G, \mathbf{y}^{(1:t)}; \theta\right) p_M\left( G| \mathbf{y}^{(1:t)}; \theta \right).
\end{equation}
\vspace{-0.2in}
\begin{figure*}[h]
    \centering
    \includegraphics[width=.8\linewidth]{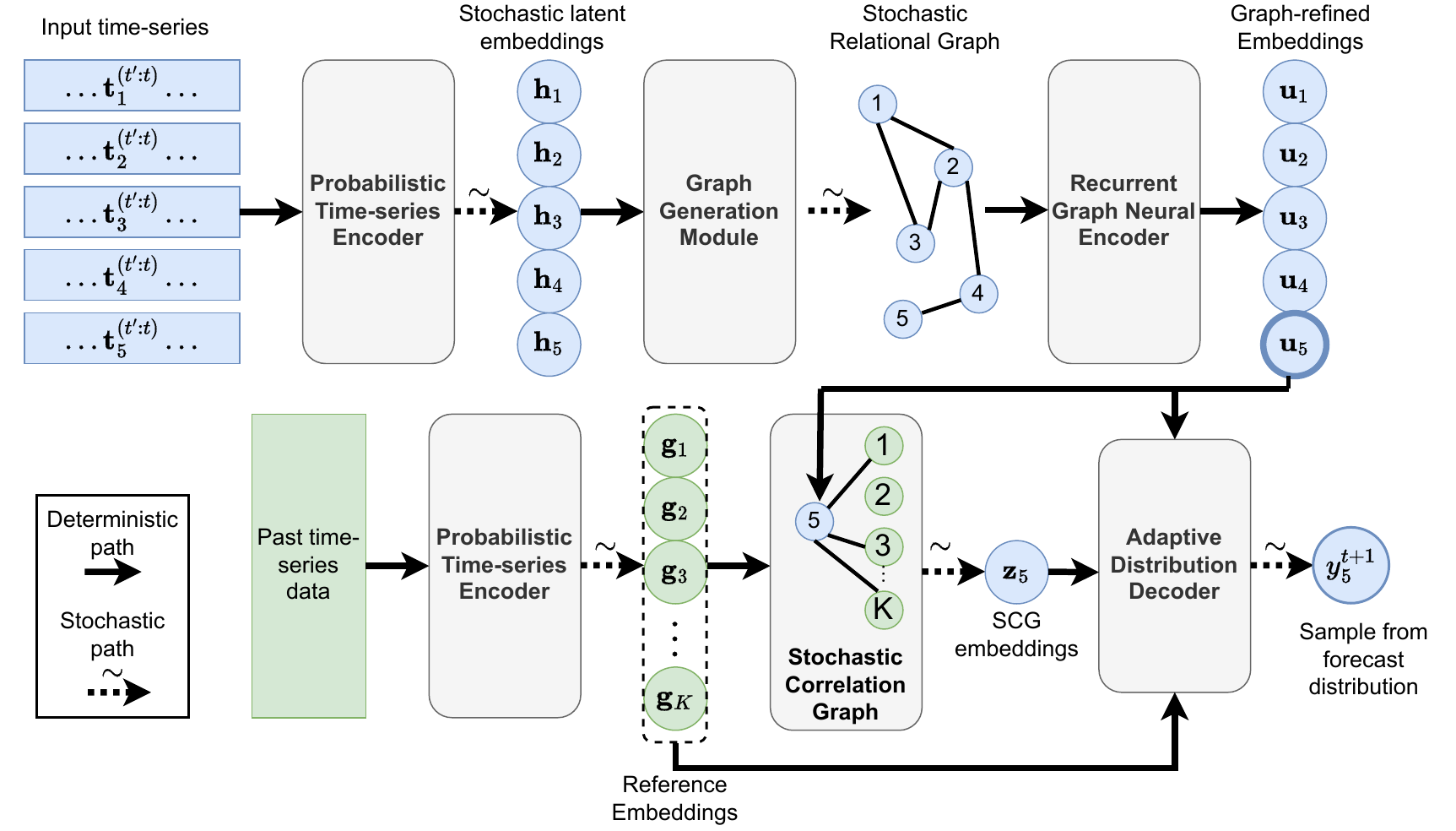}
    \caption{Overview of \model's pipeline. Probabilistic Time-series encoder encodes the input time-series into stochastic latent embeddings which are used to learn the relational graph.
        The relational graph is leveraged by Recurrent Graph Neural Encoder to model both temporal and structural patterns of multivariate time-series represented by Graph-refined embeddings. Reference Correlation Network extracts similar patterns from past data in form of RCN embeddings.
        Finally, the decoder uses Graph-refined embeddings and RCN embeddings to derive the parameters of the forecast distribution.
    }
    \label{fig:pipe}
\end{figure*}
\subsubsection{Overview}
\model models stochasticity and uncertainty of time-series when generating structural relations across time-series.
It also adaptively leverages relations and uncertainty from past data using the functional process framework \cite{louizos2019functional,kamarthi2021doubt}. \model's generative process can be summarized as:
    (1) The input time-series values are encoded using a \textit{Probabilistic Time-series Encoder} (\pte) that models a multivariate Gaussian Distribution to model each time-series capturing both time-series patterns and inherent uncertainty.
    (2) The similarity between the sampled stochastic encoding of each time-series from PTE is used to sample a graph via the \textit{Graph Generation Module} (\ggm).
    (3) Recurrent Graph Neural Encoder (\rgne) contains a series of Recurrent neural layers and Graph Convolutional Networks which derive the encoding of each time-series leveraging the learned graph.
    (4) We also model the similarity of encodings of input time-series with past data using a reference correlation network (\scg).
    (5) Finally, \model uses the graph-refined embeddings and  historical information from \scg to learn the parameters of the output distribution.

\subsubsection{Probabilistic Time-series Encoder}
We first model both the information and uncertainty for each of the $N$ time-series, by using deep sequential models to capture complex temporal patterns of the input time-series sequence $\mathbf{y}_i^{(t':t)}$.
We use a GRU \cite{cho2014properties} followed by \textit{Self-attention} \cite{vaswani2017attention} over the hidden embeddings of GRU:
\begin{equation}
    \{\mathbf{\bar{h}_i}\}_{i=1}^N = \{\text{Self-Atten}(GRU(\mathbf{y}_i^{t':t}))\}_{i=1}^N.
\end{equation}
We then model the final latent embeddings of univariate time-series as a multivariate Gaussian distribution similar to VAEs \cite{kingma2019introduction}:
\begin{equation}
    \mu_{\mathbf{h}_i}, \log\sigma_{\mathbf{h}_i} = NN_h(\mathbf{\bar{h}}_i), \quad \mathbf{h}_i \sim \mathcal{N}(\mu_{\mathbf{h}_i}, \sigma_{\mathbf{h}_i}).
\end{equation}
where $NN_h$ is a single layer of feed-forward neural network. The output latent embeddings $\mathbf{H} = \{\mathbf{h}_i\}_i$ are stochastic embeddings sampled from Gaussian parameterized by $\{\mu(\mathbf{h}_i), \sigma(\mathbf{h}_i)\}_i$. which captures uncertainty of temporal patterns.

\subsubsection{Probabilistic Graph Generation Module}
Since it is computationally expensive to model all possible relations between time-series, we aim to mine sparse stochastic relations between time-series along with the uncertainty of the underlying relations.
We generate a stochastic relational graph (SRG) $G$ of $N$ nodes that model the similarity across all time-series.
We use a stochastic process to generate $G$ leveraging stochastic latent embeddings $\mathbf{H}$ from \pte.
We parametrize the adjacency matrix $\mathbf{A}(\mathbf{H}), \in \{0,1\}^{N \times N}$ of $G$ by modelling existence of each edge $A_{i,j}$ as a Bernoulli distribution parameterized by $\theta_{ij}$ derived as:
\begin{equation}
    \theta_{i,j} = \text{sig}(NN_{G_2}( NN_{G_1}(\mathbf{h}_i)) + NN_{G_1}(\mathbf{h}_j))
    \label{eqn:theta}
\end{equation}
where $NN_{G1}$ and $NN_{G2}$ are feed-forward networks.
We sample the adjacency matrix which captures temporal uncertainty in $\mathbf{h}_i, \mathbf{h}_j$ and relational uncertainty in $\theta_{i,j}$.
\begin{equation}
    (A_{i,j} = A_{j,i}) \sim \text{Bernoulli}(\theta_{i,j}); \forall i \leq j.
\end{equation}

\subsubsection{Recurrent Graph Neural Encoder}
We combine the relational information from SRG with temporal information   via a combination of Graph Neural Networks and  recurrent networks:
\begin{equation}
        \mathbf{v_i^{(t)}} = \text{GRU-Cell}(\mathbf{u}_i^{(t-1)}, y_{i}^{(t)}) ;
        \{\mathbf{u}_i^{(t)}\}_i = GNN(\{\mathbf{v}_i^{(t)}\}, A)
\end{equation}
We input $h_i$ as the initial hidden embedding for GRU-Cell at initial time-step $t'$ to impart temporal information from PTE.
We finally combine the intermediate embeddings $\{\mathbf{u}_i^{(t)}\}_{t=t'}^t$ using self-attention to get Graph-refined Embeddings $\mathbf{U} = \{\mathbf{u}_i\}_{i=1}^N$ where:
\begin{equation}
    \mathbf{u}_i = \text{Self-Attention}(\{\mathbf{u}_i^{(t)}\}_t).
\end{equation}

\subsubsection{Reference Correlation Network}
This historical similarity is useful since time-series shows similar patterns to past data.
Therefore, we model relations with past historical data of all $N$ time-series of datasets.
We  encode the past information of all time-series into \textit{reference embeddings} $\mathbf{G} = \{\mathbf{g}_j\}_{j=1}^K$:
\begin{equation}
    \mu_{\mathbf{g}_j}, \sigma_{\mathbf{g}_j} = \pte(\mathbf{y}_j^{(1:t)}), \quad \mathbf{g}_j \sim \mathcal{N}(\mu_{\mathbf{g}_j}, \sigma_{\mathbf{g}_j}).
\end{equation}
Then, similar to \cite{kamarthi2021doubt} we sample edges of a bipartite Reference Correlation Network $S$ between \textit{reference embeddings} $\mathbf{G}$ and Graph-refined Embeddings $\mathbf{U}$ based on their similarity as:
\begin{equation}
    \kappa(\mathbf{u}_i,\mathbf{g}_j) = \exp(-\gamma||\mathbf{u}_i-\mathbf{g}_j||^2), \quad S_{i,j} \sim \text{Bernoulli} (\kappa(\mathbf{u}_i,\mathbf{g}_j)).
\end{equation}
where $\gamma$ is a learnable parameter.
To leverage the similar reference embeddings sampled for each time-series $i$, we aggregate the sampled reference embeddings to form the RCN embeddings $\mathbf{Z} = \{\mathbf{z}_i\}_{i=1}^N$ as:
\begin{equation}
    \mathbf{z}_i \sim \mathcal{N}\left( \sum_{j: S_{ij} = 1} NN_{z1}(\mathbf{g}_j), \exp(\sum_{j: S_{ij} = 1} NN_{z2}(\mathbf{g}_j)) \right)
\end{equation}
where $NN_{z1}$ and $NN_{z2}$ are single fully-connected layers.
Therefore, the data from reference embeddings that show similar patterns to input time-series are more likely to be sampled.

\subsubsection{Adaptive Distribution Decoder}
The decoder parameterizes the forecast distribution using multiple perspectives of information and uncertainty from previous modules:
 Graph-refined embeddings $\mathbf{U}$, RCN embeddings $\mathbf{Z}$ and a global embedding of all reference embeddings $g$ derived as:
$
    \mathbf{g} = \text{Self-Attention}(\{\mathbf{g}_j\}_j).
$
However, information from each of the modules may have varied importance based.
Therefore, we use a weighted aggregation of these embeddings:
$
    \mathbf{k}_i = l_1\mathbf{u}_i + l_2\mathbf{z}_i + l_3 \mathbf{g}
    \label{eqn:decoderwt}
$
to get the input embedding $\mathbf{k}_i$ for the decoder where $\{l_i\}_{i=1}^3$ are learnable parameters.
The final output forecast distribution is derived as:
\begin{equation}
    y_{i}^{(t+1)} \sim \mathcal{N}(NN_{y1}(\mathbf{k}_i), \exp(NN_{y2}(\mathbf{k}_i)))
\end{equation}

\subsubsection{Training and Inference}
The full generative pipeline of \model is:
\[
    \begin{split}
        &P(\mathbf{y^{(t+1)}}| \mathbf{y}^{t':t}, \mathcal{D}) = \int \underbrace{P(\mathbf{H}|\mathbf{y}^{t':t})}_{\text{Time-series Encoder} (\pte)}\\
       &\underbrace{P(A|\mathbf{H})}_{\text{Graph Generation}(\ggm)}
        \underbrace{P(\mathbf{U}|\mathbf{H},A)}_{\text{Recurrent Graph Neural Encoder}(\rgne)}\\
        &\underbrace{P(S, \{\mathbf{g}_j\}_j|\mathbf{U}, \mathcal{D}) P(\mathbf{Z}|S,\{\mathbf{g}_j\}_j) }_{\text{Reference Correlation Network}(\scg)}
        \underbrace{P(\mathbf{y^{(t+1)}} | \mathbf{Z, U, g})}_{\text{Decoder}}
        d\mathbf{H} dA d\mathbf{U} d\mathbf{g} dS d\mathbf{Z}.
    \end{split}
\]
We train the parameters of \model to increase the log-likelihood loss $\log P(\mathbf{y^{(t+1)}}| \mathbf{y}^{t':t}, \mathcal{D})$.
Since integration over high-dimensional latent random variables is intractable,
we use amortized variational inference like \cite{louizos2019functional} and construct the variation distribution
$q(\mathbf{H,U,Z},S,A | \mathbf{y}^{t':t}, \mathcal{D}) = P(\mathbf{H}|\mathbf{y}^{t':t}) P(A|\mathbf{H}) P(S|\mathbf{U}, \mathcal{D}) q_1(\mathbf{U}, \mathbf{Z} | \mathbf{y}^{t':t})$
where $q_1$ is a fully connected network over $\{\bar{\mathbf{h}}_i\}$ that parameterizes the variational distributions for $\mathbf{U}$ and $\mathbf{Z}$.
The loss is optimized using stochastic gradient descent based Adam optimizer~\cite{kingma2014adam}.
During inference, we generate Monte Carlo samples from the full distribution $P(\mathbf{y^{(t+1)}}| \mathbf{y}^{t':t}, \mathcal{D})$ with discrete sampling.

\section{Experiment Setup}

\subsubsection{Baselines}
We compare with general state-of-art forecasting models that include 
(1) statistical models like ARIMA~\cite{hyndman2018forecasting},
(2) general forecasting models: \deepar \cite{salinas2020deepar},
\tsfnp \cite{kamarthi2021doubt}
(3) Graph-learning based forecasting models: \mtgnn \cite{wu2020connecting},
\gdn \cite{deng2021graph}, \gts \cite{shang2021discrete}, \nir \cite{kipf2018neural}.
Note that we are performing probabilistic forecasting while most of the baselines are modeled for point forecasts. Therefore, for methods like ARIMA, \deepar, \mtgnn and \gdn, we leverage an ensemble of ten independently trained models to
derive the forecast distribution \cite{lakshminarayanan2017simple}. 

\subsubsection{Datasets}
We evaluate our models against eight multivariate time-series datasets
from a wide range of applications that have been used in past works.
The main statistics of the datasets are summarized in Table \ref{tab:data}.
(1) \flu: We forecast a symptomatic measure of flu incidence rate based on wILI (weighted Influenza-like illness outpatients) that are provided by CDC for each of the 50 US states.
          We train on seasons from 2010 to 2015 and evaluate on seasons from 2015 to 2020.
(2) \covid: We forecast the weekly incidence of Covid-19 mortality from June 2020 to March 2021 for each of the 50 US states \cite{rodriguez2021deepcovid,kamarthi2022camul} using incidence data starting from April 2020.
(3) Similar to \cite{pamfil2020dynotears}, we use the daily closing prices for stocks of companies in S\& P 100 using the \texttt{yfinance} package \cite{yfinance72:online} from July 2014 to June 2019. The last 400 trading days are used for testing.
(4) \power: We use a popular multivariate time-series dataset for power consumption forecasting used in past works \cite{zugner2021study}.
We forecast power consumption for 15-60 minutes.
          We train for 1 year and test on data from the next year.
(5) \textit{Traffic prediction:} We use 2 datasets related to traffic speed prediction.
          \trafficla and \trafficbay \cite{li2018diffusion} are datasets of traffic speed at various spots in Los Angeles and San Francisco. We use the last 10\% of the time-series for testing.
(6) \textit{Transit demand:}
          \nycb and \nyct \cite{ye2021coupled} measure bike sharing and taxi demand respectively in New York from April to June 2016. 
\begin{table}[h]
\centering
\caption{Statistics of benchmark datasets}
\vspace{-0.1in}
\scalebox{0.85}{
\begin{tabular}{c|cccc}
Dataset      & Nodes & \# time-steps & Sample rate & \# Forecast horizon \\ \hline
\flu & 50    & 544           & 1 week      & 4 weeks             \\
\covid     & 50    & 44            & 1 week      & 4 weeks             \\
\stocks     & 100   & 1257          & 1 day       & 5 days              \\
\power  & 321   & 140256        & 15 min.     & 60 min.             \\
\trafficla      & 207   & 34272         & 5 min.      & 60 min.             \\
\trafficbay    & 325   & 52116         & 5 min.      & 60 min.             \\
\nycb     & 250   & 4368          & 30 min.     & 120 min.            \\
\nyct     & 266   & 4368          & 30 min.     & 120 min.           
\end{tabular}
}
\vspace{-0.1in}
\label{tab:data}
\end{table}

\subsubsection{Evaluation Metrics}
We evaluate our model and baselines using carefully chosen metrics that are widely used in forecasting literature.\footnote{Code: \url{https://anonymous.4open.science/r/Stoic_KDD24-D5A8}. Supplementary: \url{https://anonymous.4open.science/r/Stoic_KDD24-D5A8/Struct_Supp.pdf}}
They evaluate for both accuracy of the mean of forecast distributions as well as calibration of the distributions \cite{hyndman2006another,tsyplakov2013evaluation,jordan2017evaluating}.
We use RMSE for point-prediction and CRPS as well as Confidence score~\cite{kamarthi2021doubt,kuleshov2018accurate,fisch2022calibrated} for
measuring forecast calibrations.


\subsubsection{Forecast Accuracy and Calibration}
\label{sec:results}
\begin{table*}[!ht]
\centering
\caption{Average Evaluation Scores (over 20 runs) for \model and baselines. The best-performing scores are in bold and the second-best is underlined. \model provides around 9\%, 14.7\%, and 16\% better performance in terms of RMSE, CS and CRPS respectively.}
\vspace{-0.2in}
\scalebox{0.99}{
\begin{tabular}{c|ccc|ccc|ccc|ccc}
             & \multicolumn{3}{c}{\flu}              & \multicolumn{3}{c}{\covid}                  & \multicolumn{3}{c}{\stocks}                     & \multicolumn{3}{c}{\power}                 \\ \hline
             & RMSE          & CRPS          & CS            & RMSE          & CRPS          & CS            & RMSE          & CRPS          & CS            & RMSE           & CRPS           & CS            \\ \hline
ARIMA       & 0.85          & 0.91          & 0.22          & 91.5          & 87.4          & 0.21          & 0.42          & 0.40          & 0.27          & 395.2          & 417.2          & 0.37          \\
\deepar       & 0.68          & 0.97          & 0.15          & 49.1          & 48.5          & \textbf{0.19}          & 0.35          & 0.41          & 0.18          & 249.9          & 267.8          & 0.18          \\
\tsfnp        & 0.64          & \underline{0.52}          & \textbf{0.05}          & 43.6          & 44.5          & 0.15          & 0.22          & 0.28          & \underline{0.16}          & 257.3          & 277.0          & 0.15          \\
MTGNN        & 0.68          & 0.75          & 0.18          & 48.3          & 49.5          & 0.22          & 0.17          & 0.26          & 0.19          & \textbf{189.3} & \textbf{207.1} & 0\underline{0.13}          \\
GDN          & 0.73          & 0.71          & 0.15          & 42.74         & 43.2          & 0.25          & 0.18          & 0.24          & 0.24          & 249.1          & 263.6          & 0.21          \\
\nir          & 0.71          & 0.74          & 0.22          & 63.44         & 72.19         & 0.35          & \underline{0.14}          & 0.23          & 0.19          & 315.9          & 338.6          & 0.18          \\
GTS          & \underline{0.63}          & 0.66          & 0.17          & \underline{41.96}         & \underline{36.25}         & 0.22          & \textbf{0.11} & 0.18          & 0.22          & 197.4          & 208.0          & 0.16          \\
LDS          & 0.69          & 0.64          & 0.16          & 45.17         & 41.32         & 0.23          & 0.15          & 0.19          & \underline{0.16}          & 213.6          & 229.0          & 0.19          \\ \hline
\model         & \textbf{0.57} & \textbf{0.42} & \underline{0.07} & \textbf{32.7} & \textbf{31.7} & \underline{0.2}  & \textbf{0.11} & \textbf{0.16} & \textbf{0.12} & \underline{191.6} & \underline{207.3} & \textbf{0.09} \\ \hline \hline
             & \multicolumn{3}{c}{\trafficla}                   & \multicolumn{3}{c}{\trafficbay}                 & \multicolumn{3}{c}{\nycb}                  & \multicolumn{3}{c}{\nyct}                    \\ \hline
             & RMSE          & CRPS          & CS            & RMSE          & CRPS          & CS            & RMSE          & CRPS          & CS            & RMSE           & CRPS           & CS            \\ \hline
ARIMA       & 2.17         & 2.06          & 0.23          & 4.1           & 3.9           & 0.14          & 4.27          & 4.11          & 0.34          & 18.55          & 17.49          & 0.15          \\
\deepar       & 2.05          & 1.93          & 0.17          & 3.8           & 4.2           & 0.16          & 3.11          & 3.15          & 0.28          & 14.52          & 15.77          & 0.17          \\
\tsfnp        & 2.11          & 1.89          & 0.14          & 4.1           & 3.9           & 0.13          & 2.94          & 2.85          & 0.23          & 15.13          & 14.97          & 0.15          \\
MTGNN        & \underline{1.86}          & 1.64          & 0.14          & \underline{3.2}           & \underline{2.9}           & \underline{0.11}          & 2.62          & \underline{2.61}          & \underline{0.17}          & \underline{10.37}          & \underline{10.13}          & \underline{0.13}          \\
GDN          & 1.95          & 1.83          & 0.16          & 3.4           & 3.1           & 0.16          & 2.74          & 1.78          & \textbf{0.15}          & 11.15          & 10.64          & 0.17          \\
\nir          & 2.17          & 1.95          & 0.19          & 7.9           & 5.3           & 0.14          & 3.18          & 4.7           & 0.36          & 14.38          & 17.52          & 0.31          \\
GTS          & 1.88          & \underline{1.74}          & 0.15          & 3.7           & 3.2           & 0.13          & 2.76          & 2.85          & 0.24          & 10.75          & 12.68          & 0.18          \\
LDS          & 2.12          & 2.05          & \underline{0.13}          & 3.5           & 3.7           & 0.18          & \underline{2.55}          & 2.67          & 0.21          & 11.44          & 13.98          & 0.19          \\ \hline
\model         & \textbf{1.84} & \textbf{1.48} & \textbf{0.11} & \textbf{2.8}  & \textbf{2.5}  & \textbf{0.08} & \textbf{2.52} & \textbf{2.41} & \textbf{0.15} & \textbf{8.43}  & \textbf{8.61}  & \textbf{0.09}
\end{tabular}
}
\label{tab:forecast}
\end{table*}
We evaluate the average performance of \model and all the baselines across 20 independent runs in Table \ref{tab:forecast}.
\model provides 8.8\% more accurate forecasts (RMSE scores) across all benchmarks
with an impressive 16.5\% more accurate forecasts in Epidemic forecasting (\flu and \covid) and is 9\% better in traffic benchmarks.
For calibration,
we observe significantly better CS scores across all benchmarks.
\model also provides
16\% higher CRPS scores over the best-performing baseline in each task. \model has 34.5\% and 11.5\% better performance in epidemic forecasting and traffic forecasting respectively.
In particular, we also observe that baselines like ARIMA, \deepar, and \tsfnp which do not learn a graph have 7-15\% poorer performance in traffic benchmarks and over 180\% poorer performance in \stocks benchmark compared to other baselines that learn a graph. 

\subsubsection{Robustness to Noise}
\label{sec:robustness}
We evaluate the efficacy of modeling the underlying structure of time-series and learning
uncertainty of each time-series  in
helping models adapt to noise in the datasets.
Learning a single global structure to model relations across time-series 
as well as modeling uncertainty in data can help the
model to adapt to noise in datasets
during inference.
We therefore expect \model to be further resilient to noise in data during testing.
We use the models trained on clean training time-series datasets and inject noise to
time-series input during testing.
Each input time-series is first independently normalized with 0 mean and unit standard deviation
and then add a gaussian noise with standard deviation $\rho$.
\begin{figure*}[h]
    \centering
    \begin{subfigure}[b]{0.166\linewidth}
        \centering
        \includegraphics[width=.9\linewidth]{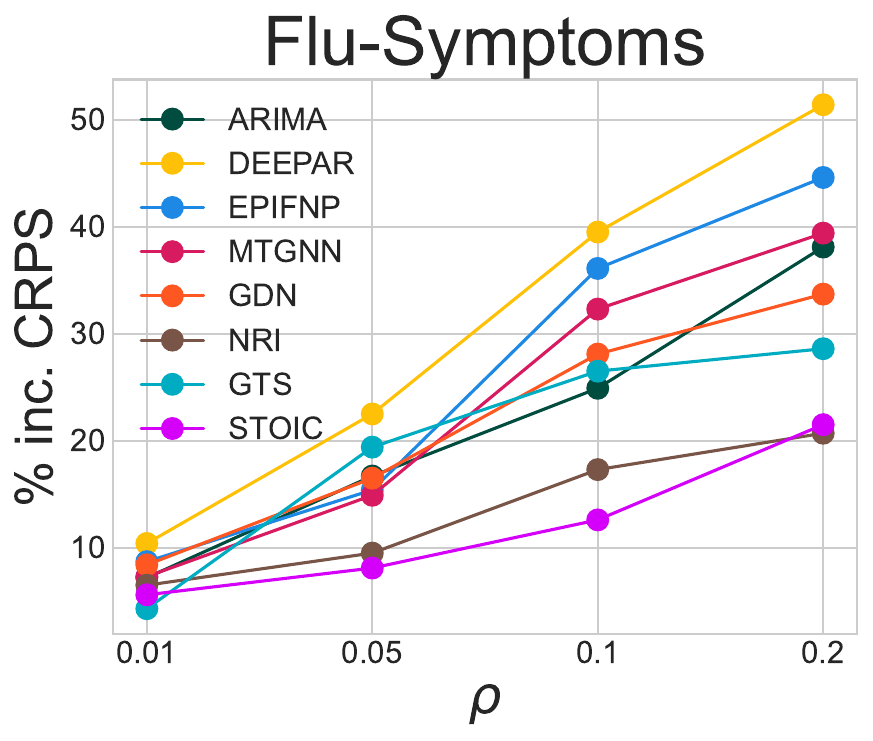}
    \end{subfigure}%
    \begin{subfigure}[b]{0.166\linewidth}
        \centering
        \includegraphics[width=.9\linewidth]{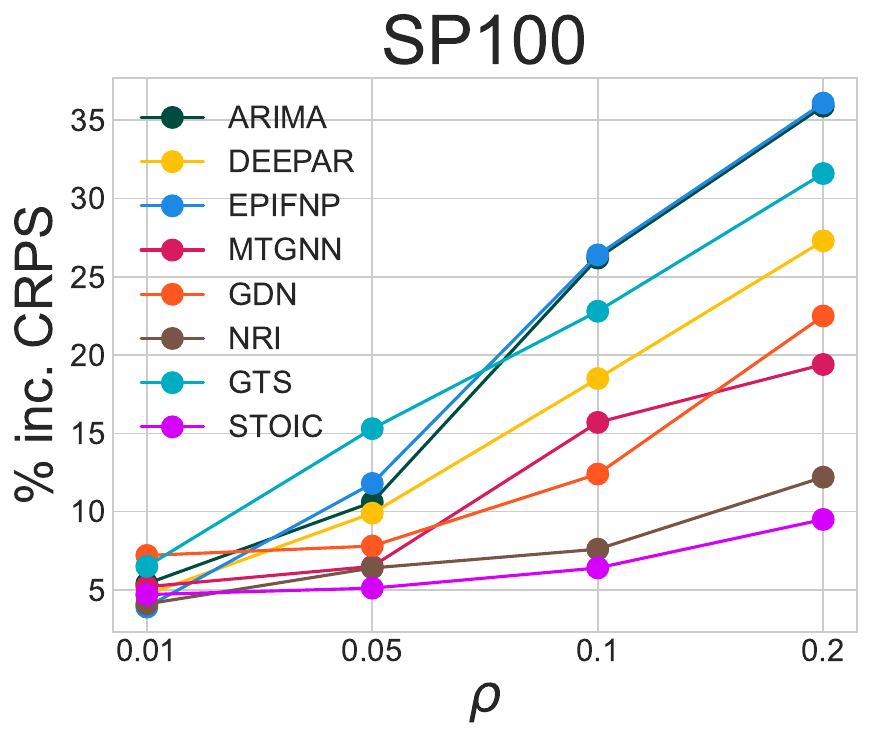}
    \end{subfigure}%
    \begin{subfigure}[b]{0.166\linewidth}
        \centering
        \includegraphics[width=.9\linewidth]{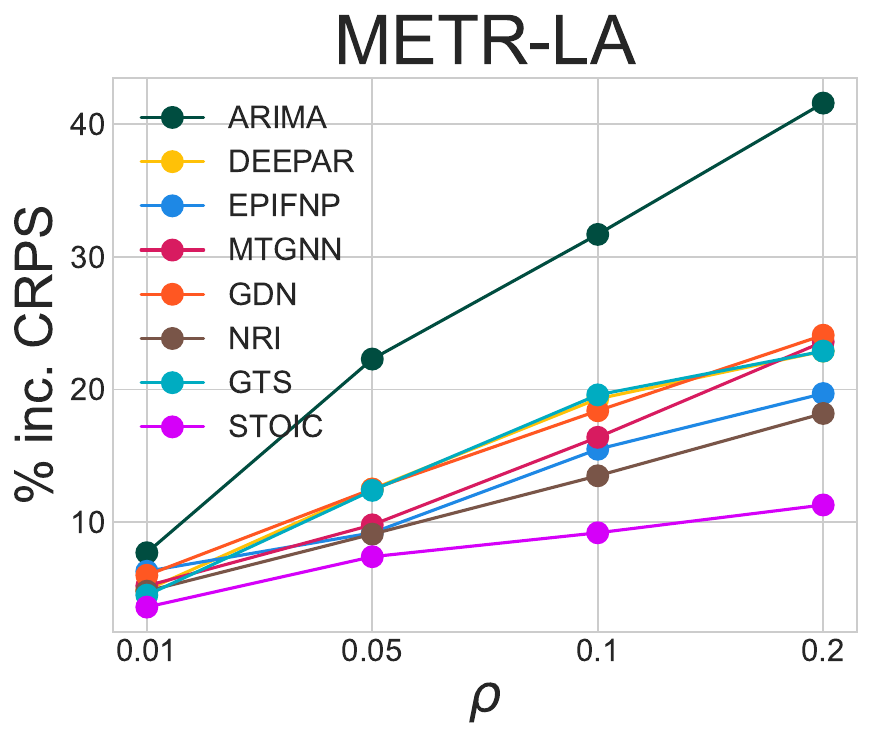}
    \end{subfigure}%
    \begin{subfigure}[b]{0.166\linewidth}
        \centering
        \includegraphics[width=.9\linewidth]{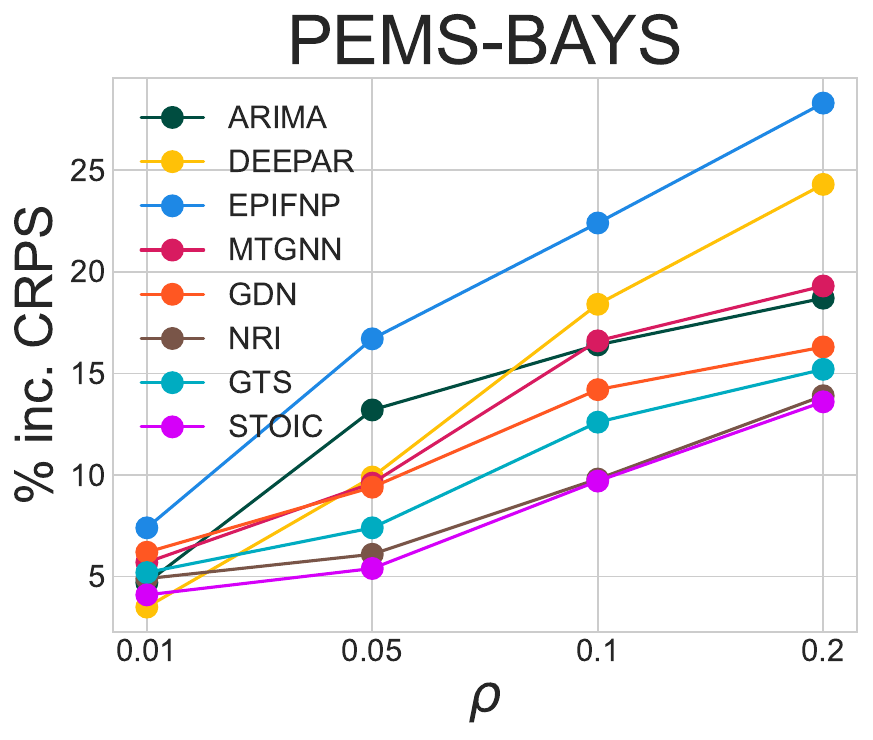}
    \end{subfigure}%
    \begin{subfigure}[b]{0.166\linewidth}
        \centering
        \includegraphics[width=.9\linewidth]{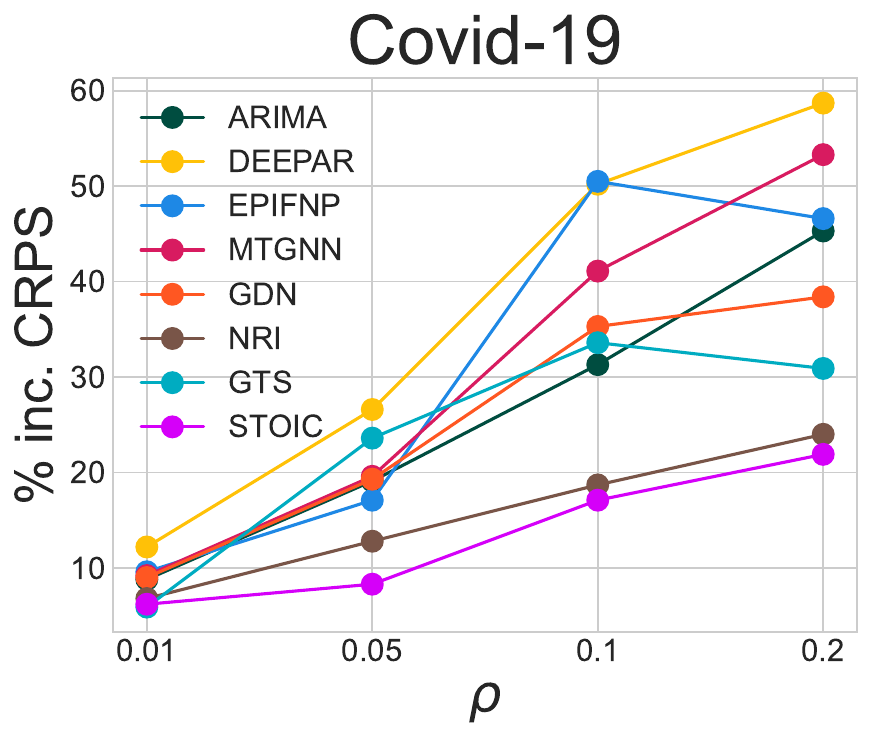}
    \end{subfigure}%
    \begin{subfigure}[b]{0.166\linewidth}
        \centering
        \includegraphics[width=.9\linewidth]{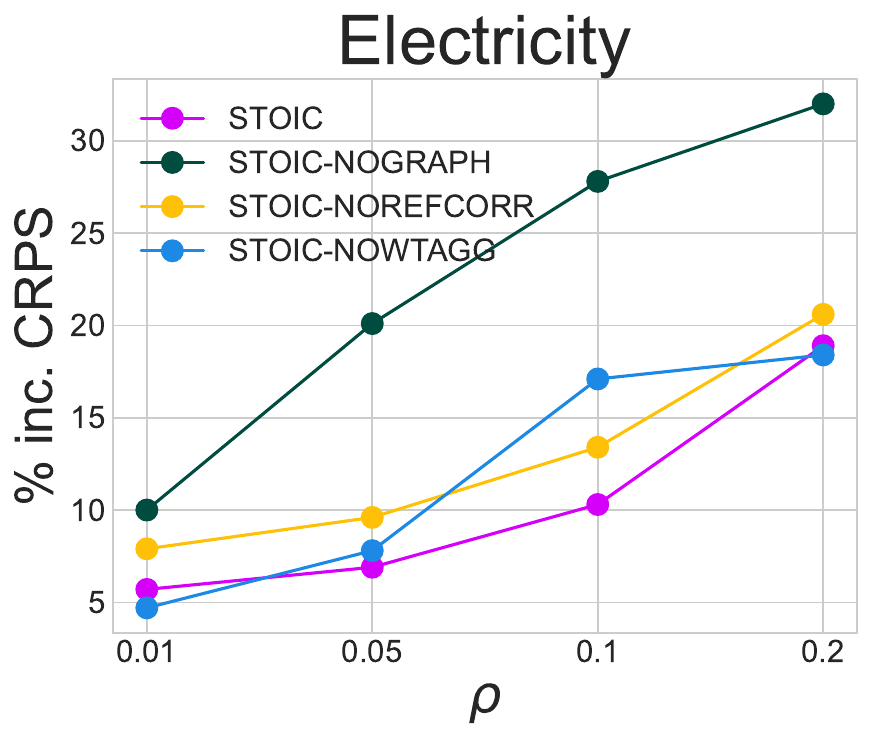}
    \end{subfigure}%
    \caption{\% increase in CRPS with an increase in gaussian Noise ($\rho$) added to the input sequence. Performance decrease of \model is significantly lower than most of the baselines.}
    \label{fig:noise_main}
\end{figure*}

We plot the decrease in performance (measured using CRPS) with an increase in noise, measured by $\rho$ in Figure \ref{fig:noise_main}.
First, we observe that as we increase $\rho$ the performance decreases for all models.
We observe that the baselines that do not learn a graph structure
on average show 45-60\% larger decrease in performance compared to the rest of the baselines
showing the efficacy of structure learning for robust forecasting.
However, \model's decrease in performance is significantly less compared to most of the baselines.
At $\rho=0.2$ the average performance decrease in all baselines is at 19\% - 27\%
whereas \model's performance decrease is at 9-20\%.
Therefore, \model's ability to model uncertainty of time-series as well as effectively capture structural patterns
enables it to provide forecasts that are robust to noise.

\subsubsection{Ablation Studies}
We evaluate the efficacy of various modeling choices of \model.
Specifically, we access the influence of graph generation, \scg, and weighted aggregation (Equation \ref{eqn:decoderwt}) via ablation studies.
\model on average outperforms all the ablation variants with
8.5\% better accuracy and 3.2\% better CS.
Graph generation is the most impactful for performance followed by \scg.
For additional details see Appendix~\ref{sec:ablation}.

\subsubsection{Relations Captured by Inferred Graphs}
We consider 
various meaningful domain-specific relations for 
the time-series of these datasets and study how well \model's inferred graphs capture them.
We observed that \model inferred strong relations between time-series of stocks of the same sectors in \stocks.
In the case of \trafficbay and \trafficla, the graph inferred
by \model between traffic sensors is most highly correlated to actual proximity of sensors with each other compared to other baselines.
Finally, the graphs inferred in \flu and \covid
are correlated with geographical adjacency of regions and road density of connecting regions.
We provide more details in Appendix Section ~\ref{sec:graphgen}

\section{Conclusion}
We introduced \model, a probabilistic multivariate forecasting model that performs joint structure learning and forecasting leveraging uncertainty in time-series data to provide accurate and well-calibrated forecasts.
We observed that \model performs 8.5\% better in accuracy and 14.7\% better in calibration over multivariate time-series benchmarks from various domains.
Due to structure learning and uncertainty modeling, we also observed that \model better adapts to the injection of varying degrees of noise to data during inference with \model's performance drop being almost half of the other state-of-art baselines.
Finally, we observed that \model identifies useful patterns across time-series based on inferred graphs such as the correlation of stock from similar sectors,
location proximity of traffic sensors and geographical adjacency and mobility patterns across US states for epidemic forecasting.
While our work focuses on time-series with real values modeled as Gaussian distribution,
our method can be extended to other distributions modeling different kinds of continuous signals.
Further, \model only models a single global structure across time-series similar to most previous works.
Therefore, extending our work to learn dynamic graphs that can adapt to changing underlying time-series relations or model multiple temporal scales could be an important direction for future research.

\bibliographystyle{ACM-Reference-Format}
\bibliography{references}

\newpage
\clearpage
\appendix

\noindent\textbf{\Large Supplementary for the paper "Learning Latent Graph Structures for Accurate and Calibrated
    Time-series Forecasting"}

We run our models on an Nvidia Tesla V100 GPU and found that it takes less than 4 GB of memory for all benchmarks. The code for our model is available at an anonymized repository \url{https://anonymous.4open.science/r/Stoic_KDD24-D5A8} and will be released publicly when accepted.

\section{Related Work}
\subsubsection{Multivariate forecasting using domain-based structural data}
Deep neural networks have achieved great success in probabilistic time series forecasting. \deepar \cite{salinas2020deepar} trains an auto-regressive recurrent network to predict the parameters of the forecast distributions. Other works including
deep Markov models \cite{krishnan2017structured} and deep state space models \cite{rangapuram2018deep, li2021learning} explicitly model the transition and emission components with neural networks. Recently, \tsfnp \cite{kamarthi2021doubt}  leverages functional neural processes and achieves state-of-art performance in
epidemic forecasting.
However, all these methods treat each time series individually and suffer from limited forecasting performance.
Leveraging the relation structure among time-series to improve forecasting performance is an emerging area.
GCRN \cite{seo2018structured}, DCRNN \cite{li2018diffusion}, STGCN \cite{yu2017spatio} and T-GCN \cite{zhao2019t} adopt graph neural networks to capture the relationships among time series and provide better representations for each individual sequence.
However, these methods all assume that the ground-truth graph structure is available in advance, which is often unknown in many real world applications.
\subsubsection{Structure learning for time-series forecasting}
When the underlying structure is unknown, we need to jointly perform graph structure learning and time-series forecasting.  \mtgnn \cite{wu2020connecting} and \gdn \cite{deng2021graph} parameterize the graph as a degree-k graph to promote sparsity but their training can be difficult since the top-K operation is not differentiable.
\gts \cite{shang2021discrete} uses the Gumbel-softmax trick \cite{jang2016categorical} for differentiable structure learning and uses prior knowledge to regularize the graph structure. The graph learned by GTS is a global graph shared by
all the time series.
Therefore, it is not flexible since it cannot adjust the graph structure for changing inputs at inference time.
\nir \cite{kipf2018neural} employs a variational auto-encoder architecture and can produce different structures for different encoding inputs. It is more flexible than \gts but needs more memory to store the individual graphs.
However, as previously discussed, these works do not model the uncertainty of time-series during structure learning and forecasting and do not focus on the calibration of their forecast distribution.

\section{Training details}
The architecture of \pte and \scg is similar to \cite{kamarthi2021doubt} with GRU being bi-directional and having 60 hidden units.
$NN_h, NN_{G1}, NN_{G2}, NN_{z1}$ and $NN_{z2}$ and GRU of \rgne also have 60 hidden units.
Therefore, Graph-refined embeddings $\mathbf{U}$, \scg embeddings $\mathbf{Z}$ and global embedding $\mathbf{g}$ are 60 dimensional vectors.

We used Adam optimizer \cite{kingma2014adam} with a learning rate of 0.001. We found that using a batch size of 64 or 128 provided good performance with stable training.
We used early stopping with patience of 200 epochs to prevent overfitting.
For each of the 20 independent runs, we initialized the random seeds of all packages to 1-20.
In general, variance in performance across different random seeds was not significant for all models.

\section{Ablation Studies}
\label{sec:ablation}
We evaluate the efficacy of various modeling choices of \model.
Specifically, we access the influence of graph generation, \scg, and weighted aggregation (Equation \ref{eqn:decoderwt}) via the following ablation variants of \model:
\begin{itemize}
  \item \modelng: We remove the \ggm and GCN modules of \rgne and therefore do not use Graph-refined embeddings $\mathbf{U}$ in the decoder.
  \item \modelns: We remove the \scg module and do not use \scg embeddings $\mathbf{Z}$ in the decoder.
  \item \modelnc: We replace weighted aggregation with concatenation in Equation \ref{eqn:decoderwt} as
        \begin{equation}
          \mathbf{k}_i = \mathbf{u}_i \oplus \mathbf{z}_i \oplus \mathbf{g}
        \end{equation}
        where $\oplus$ is the concatenation operator.
\end{itemize}
\begin{table*}[h]
\centering
\caption{Average Evaluation Scores (over 20 runs) for \model and ablation variants.}
\scalebox{0.99}{
\begin{tabular}{c|ccc|ccc|ccc|ccc}
             & \multicolumn{3}{c}{\flu}              & \multicolumn{3}{c}{\covid}                  & \multicolumn{3}{c}{\stocks}                     & \multicolumn{3}{c}{\power}                 \\ \hline
             & RMSE          & CRPS          & CS            & RMSE          & CRPS          & CS            & RMSE          & CRPS          & CS            & RMSE           & CRPS           & CS            \\ \hline
\model         & \textbf{0.57} & \textbf{0.42} & \underline{0.07} & \underline{32.7} & \underline{31.7} & \underline{0.2}  & \textbf{0.11} & \textbf{0.16} & \textbf{0.12} & \textbf{191.6} & \underline{207.3} & \textbf{0.09} \\
\modelng & 0.67          & 0.51          & 0.06          & 42.7          & 41.6          & 0.11          & 0.24          & 0.31          & {0.16}          & 226.5          & 248.5          & 0.13          \\
\modelns   & 0.65          & 0.68          & 0.15          & 44.72         & 40.66         & 0.19          & \underline{0.12}          & 0.21          & \underline{0.15}          & \underline{196.4}          & \underline{211.0}          & \underline{0.11}          \\
\modelnc  & 0.63          & 0.44          & \textbf{0.06}          & \textbf{31.6} & \textbf{28.3} & \textbf{0.19} & 0.13          & \underline{0.19}          & 0.18          & 217.5          & 237.6          & 0.18          \\ \hline \hline
             & \multicolumn{3}{c}{\trafficla}                   & \multicolumn{3}{c}{\trafficbay}                 & \multicolumn{3}{c}{\nycb}                  & \multicolumn{3}{c}{\nyct}                    \\ \hline
             & RMSE          & CRPS          & CS            & RMSE          & CRPS          & CS            & RMSE          & CRPS          & CS            & RMSE           & CRPS           & CS            \\ \hline
\model         & \textbf{1.84} & \textbf{1.48} & \textbf{0.11} & \textbf{2.8}  & \textbf{2.5}  & \textbf{0.08} & \textbf{2.52} & \textbf{2.41} & \textbf{0.15} & \textbf{8.43}  & \textbf{8.61}  & \textbf{0.09} \\
\modelng & 2.19          & 1.85          & 0.17          & 3.9           & 4             & 0.15          & 3.41          & 3.16          & 0.19          & 13.65          & 12.89          & 0.21          \\
\modelns   & 2.25          & 2.18          & 0.15          & \underline{3.4}           & \underline{3.1}           & 0.12          & \underline{2.86}          & 2.72          & \underline{0.17}          & 10.53          & 9.24           & 0.14          \\
\modelnc  & \underline{1.86}          & \underline{1.54}          & \underline{0.14}          & 3.7           & 3.6           & \underline{0.09}          & 2.95          & \underline{2.69}          & 0.21          & \underline{9.25}           & \underline{8.37}           & \underline{0.11}         
\end{tabular}
}
\label{tab:forecast_abl}
\end{table*}
\subsubsection{Forecasting performance} As shown in Table \ref{tab:forecast_abl}, \model on average outperforms all the ablation variants with
8.5\% better accuracy and 3.2\% better CS.
On average, the worst performing variant is \modelng followed by \modelns and \modelnc, showing the importance of graph generation and leveraging learned relations across time-series.

\begin{figure*}[h]
    \centering
    \begin{subfigure}[b]{0.245\linewidth}
        \centering
        \includegraphics[width=.9\linewidth]{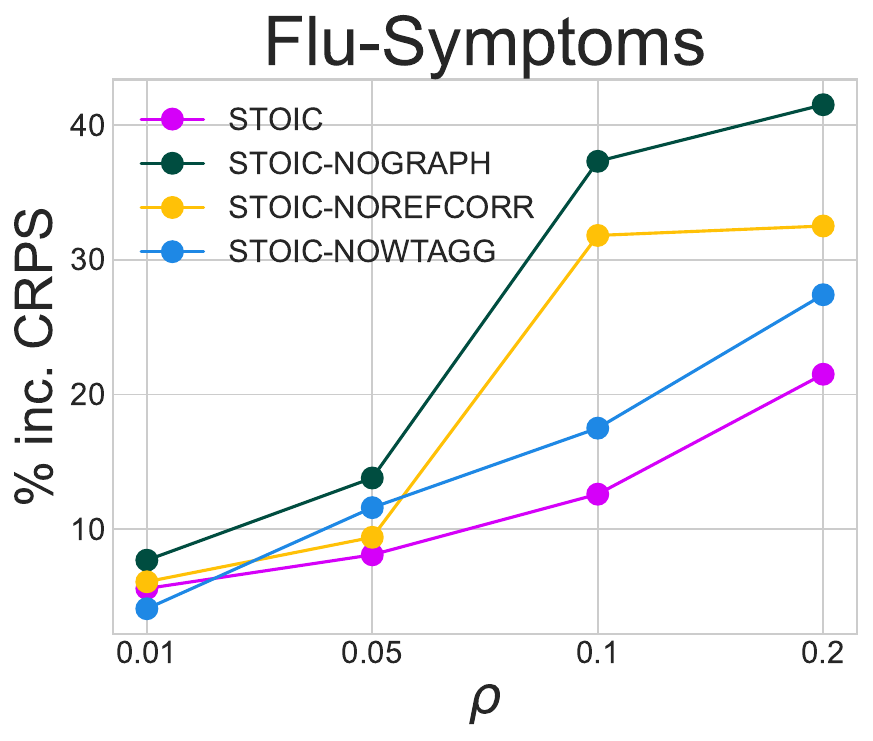}
        \caption{\flu}
    \end{subfigure}%
    \begin{subfigure}[b]{0.245\linewidth}
        \centering
        \includegraphics[width=.9\linewidth]{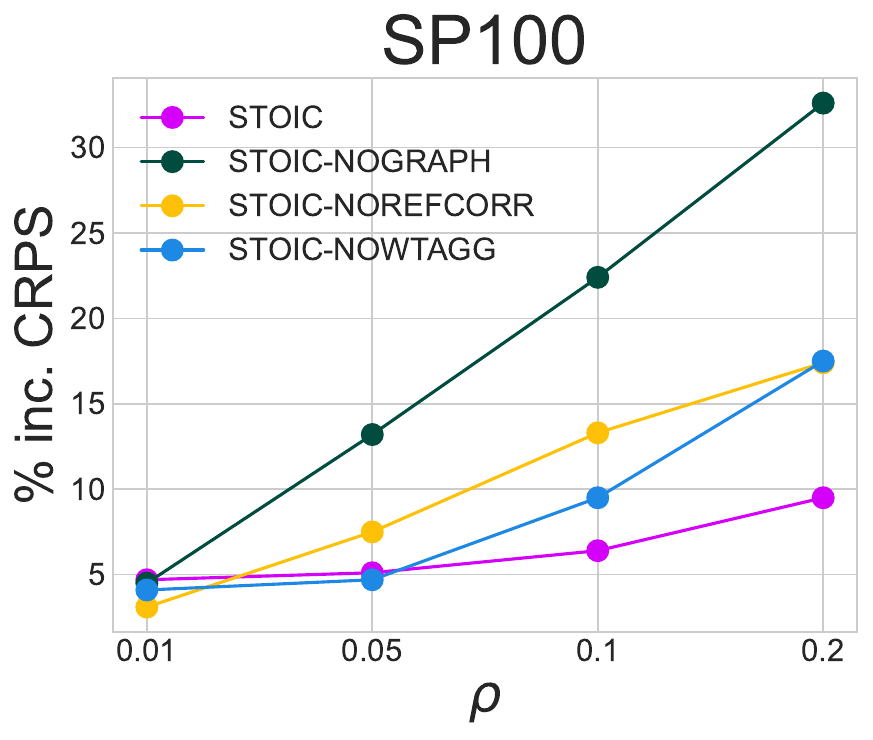}
        \caption{\stocks}
    \end{subfigure}%
    \begin{subfigure}[b]{0.245\linewidth}
        \centering
        \includegraphics[width=.9\linewidth]{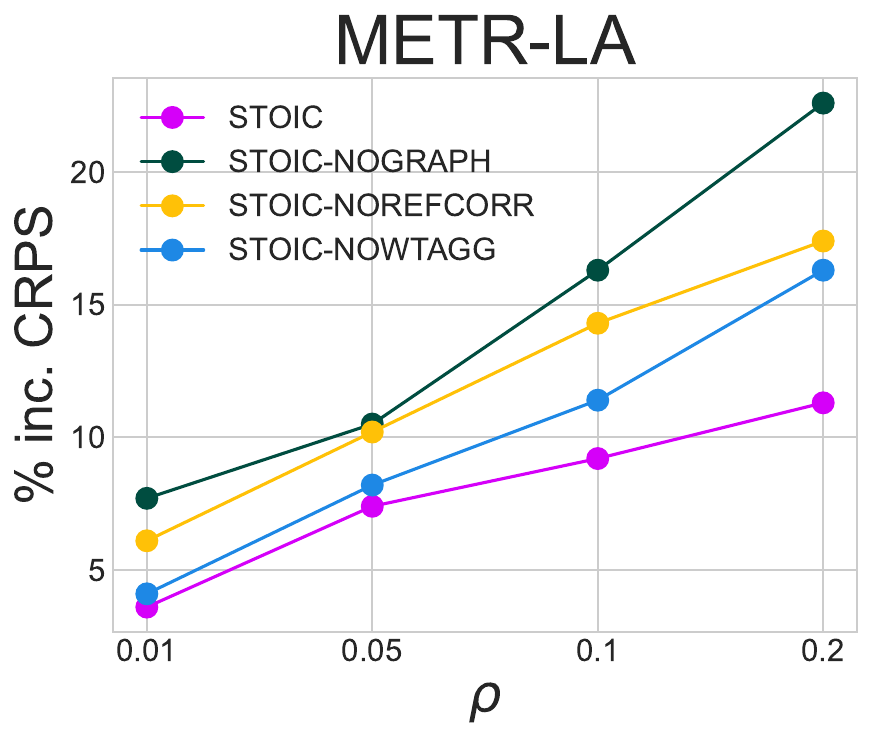}
        \caption{\trafficla}
    \end{subfigure}
    \begin{subfigure}[b]{0.245\linewidth}
        \centering
        \includegraphics[width=.9\linewidth]{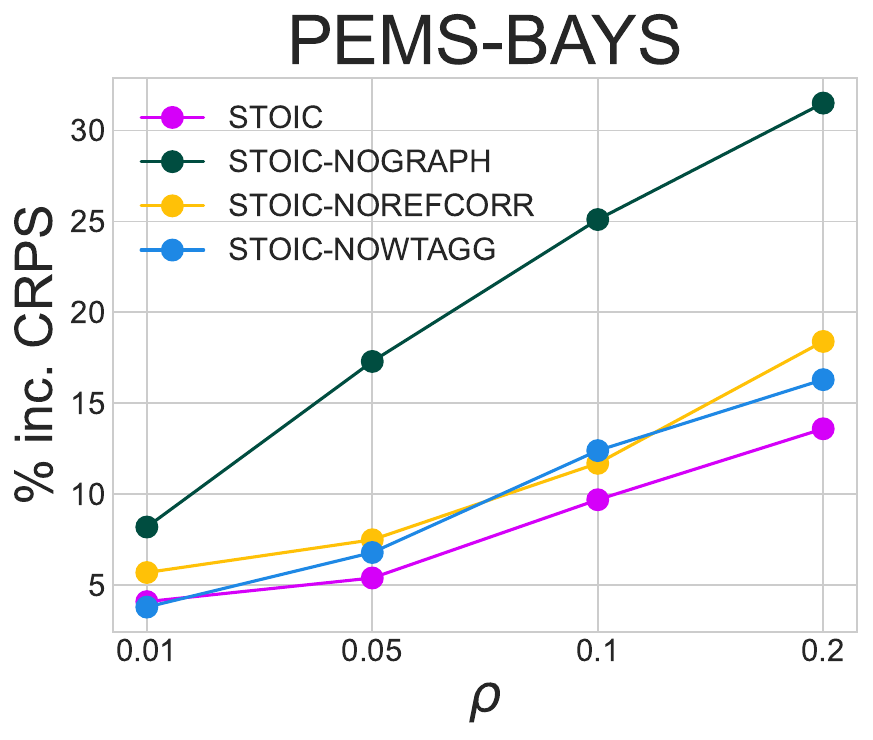}
        \caption{\trafficbay}
    \end{subfigure}%
    \begin{subfigure}[b]{0.245\linewidth}
        \centering
        \includegraphics[width=.9\linewidth]{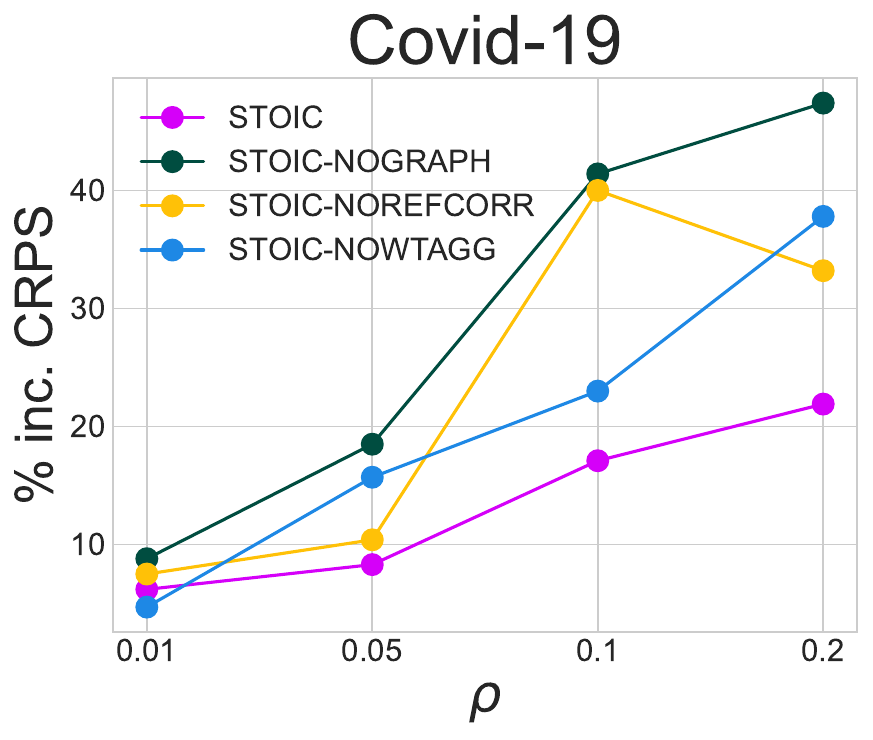}
        \caption{\covid}
    \end{subfigure}%
    \begin{subfigure}[b]{0.245\linewidth}
        \centering
        \includegraphics[width=.9\linewidth]{./Images/noise_abl_Electricity.pdf}
        \caption{\power}
    \end{subfigure}%
    \caption{\model is better adapted to noise than other ablation variants.
    \% increase in CRPS of \model and ablation variants with an increase in gaussian Noise ($\rho$) added to the input sequence.}
    \label{fig:noise_abl}
\end{figure*}
\subsubsection{Robustness to noise}
Similar to Section \ref{sec:robustness}, we also test the ablation variants' robustness in the NFI task with performance decrease compared in Figure \ref{fig:noise_abl}.
Performance continuously decreases with an increase in $\rho$.
\modelng is most susceptible to a decrease in performance with a 22-41\% decrease in
performance at $\rho=0.2$ which again shows the importance of structure to robustness.
\modelns and \modelnc's performance degradation range from 15-32\%.
Finally, we observe that \model is again significantly more resilient to noise compared to all other variants.

\subsection{Relations Captured by Inferred Graphs}
\label{sec:graphgen}
As mentioned before, for all benchmarks there is no `ground-truth' structure to compare against. 
However, in line with previous works~\cite{shang2021discrete,pamfil2020dynotears,wu2020connecting,rosensteel2021characterizing}, we consider 
various meaningful domain-specific relations for 
the time-series of these datasets and study how well \model's inferred graphs capture them through following case studies.
Note that, for \model and baselines that use a sampling strategy to construct the graph (\gts, \nir, \gts, \lds), we calculate edge probability of each pair of nodes based on sampled graphs. For other graph generating baselines, we directly use the graphs inferred.

\subsubsection{\textbf{Case Study 1: }Sector-level correlations of stocks in \stocks}
Two stocks representing companies from the same sectors typically influence each other more than stocks from different sectors.
Therefore, we measure the correlation of edge probabilities for stocks in the same sectors.
We first construct a \textit{Sector-partion graph} as follows.
We partition the stock time-series into sectors and construct a set of fully connected components where each component contains nodes of a sector. There are no edges across different sectors.
We then measure the correlation of the edge probability matrix with the adjacency matrix of 
\textit{Sector-partion graph}.

We observed a strong correlation score of 0.73 for graphs generated by \model. This was followed by graphs from \gts and \lds with correlation scores of 0.67 and 0.55.
Other baselines' graphs provided poor correlation scores below 0.35.
Interestingly, this trend also correlated with the performance in forecasting with \model, \gts and \lds being the top three best-performing models.
We also observed that the correlation score of \modelns was similar to \model whereas \modelnc also provided low correlation scores (0.39) though their forecasting performances are comparable.
\begin{table}[h]
\centering
\caption{Average edge correlation of generated graphs from the models with Sector-partition graph (for \stocks) and proximity graph (for \trafficla and \trafficbay).}
\label{tab:sp100}
\begin{tabular}{c|cccccc}
          & \multicolumn{1}{l}{MTGNN} & \multicolumn{1}{l}{GDN} & \multicolumn{1}{l}{NIR} & \multicolumn{1}{l}{GTS} & \multicolumn{1}{l}{LDS} & \multicolumn{1}{l}{\model} \\ \hline
\stocks     & 0.05                      & 0.42                    & 0.34                    & 0.67                    & 0.55                    & \textbf{0.73}            \\
\trafficla   & 0.01                      & 0.23                    & 0.16                    & \textbf{0.27}           & 0.19                    & 0.23                     \\
\trafficbay & 0                         & 0.28                    & 0.21                    & \textbf{0.7}            & 0.25                    & 0.61                    
\end{tabular}
\end{table}

\subsubsection{\textbf{Case Study 2:} Identifying location proximity in traffic forecasting benchmarks}
 Since sensors that are located close to together may have a larger probability of showing similar or correlated signals, we study if the generated graphs capture information about the proximity of the location of the traffic sensors.
We use the road location information of time-series in \trafficla and \trafficbay datasets and construct the \textit{proximity graph} based on pairwise road distances similar to \cite{zugner2021study}. Note that we did not feed any location-based information to the models during training.

We again measure the similarity of generated graphs with the proximity graph. We observe that \gts and \model provide the strongest correlation scores for both \trafficbay and \trafficla datasets with average scores of 0.7 and 0.61 for \trafficbay and 0.27 and 0.23 for \trafficla.
Due to the lower correlation scores for \trafficla, the proximity of sensors may not be useful in modeling relations across time-series.
Comparing with ablation variants, we observed that both \modelns and \modelnc showed similar correlation scores to \model.

\subsubsection{\textbf{Case Study 3:} Inferring geographical adjacency and mobility  for Epidemic Forecasting}

\begin{figure}[h]
    \centering
    \begin{subfigure}[b]{0.49\linewidth}
    \centering
    \includegraphics[width=.9\linewidth]{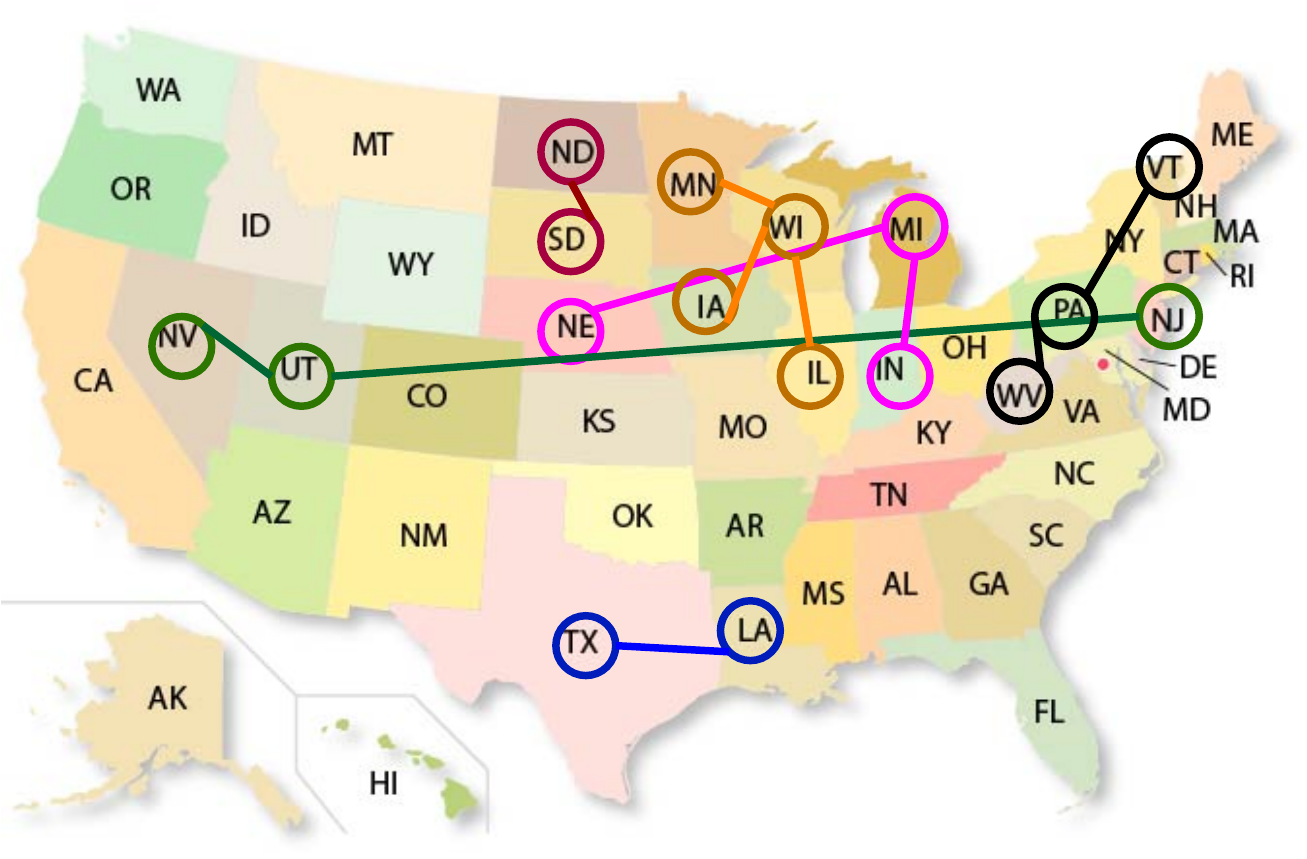}
    \caption{\covid}
    \end{subfigure}%
    \begin{subfigure}[b]{0.49\linewidth}
    \centering
    \includegraphics[width=.9\linewidth]{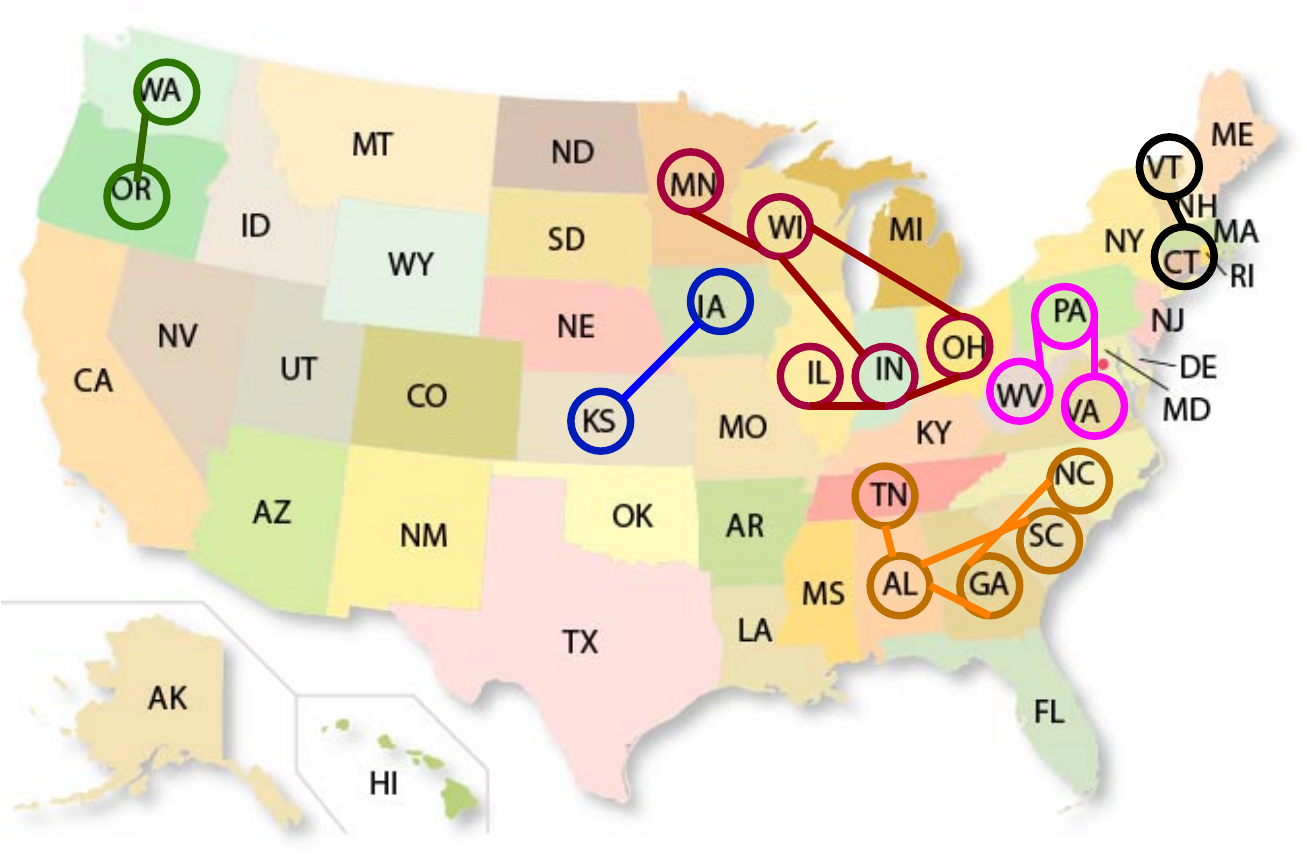}
    \caption{\flu}
    \end{subfigure}
    \caption{Most confident edges ($\theta > 0.8$) of state relation graph generated by \model for Epidemic Forecasting tasks. Note that most edges connect US states which have geographical proximity.}
    \vspace{-0.2in}
    \label{fig:epimaps}
\end{figure}

We observe the most confident edges generated by \model for \covid and \flu tasks and find that most of the edges map to adjacent states or states with strong geographical proximity similar to past works \cite{rosensteel2021characterizing}. 
Further, we observed that the specific states on which the graph relations are most confident are also connected by a higher density of roads with frequent commutes  \cite{nelson2016economic}.
This shows that \model can go beyond simple patterns and infer complex mobility patterns across states leveraging past epidemic incidence data.
Hence, \model exploits the useful relations pertaining to both geographical adjacency and mobility across these states to provide
state-of-art forecasting performance.

\end{document}